\documentclass[10pt,letterpaper,twocolumn]{article}
\usepackage[T1]{fontenc}
\usepackage{graphicx,booktabs,tabularx,array,url,color,hyperref}

\definecolor{arya}{rgb}{0.08,0.36,0.36}
\definecolor{muted}{rgb}{0.35,0.39,0.42}
\definecolor{rulegray}{rgb}{0.75,0.78,0.79}
\hypersetup{colorlinks=true,linkcolor=arya,citecolor=arya,urlcolor=arya}
\makeatletter
\renewcommand\section{\@startsection{section}{1}{\z@}{-2.6ex plus -0.6ex minus -0.2ex}{1.1ex plus 0.2ex}{\normalfont\large\bfseries\raggedright}}
\renewcommand\subsection{\@startsection{subsection}{2}{\z@}{-2ex plus -0.4ex minus -0.2ex}{0.7ex plus 0.2ex}{\normalfont\normalsize\bfseries\raggedright}}
\long\def\@makecaption#1#2{\vskip7pt{\footnotesize\noindent\textbf{#1}\enspace #2\par}\vskip4pt}
\def\ps@research{\def\@oddhead{\vbox{\hbox to\textwidth{\footnotesize\color{muted}Permission to Retry\hfill Arya.ai}\vskip6pt{\color{rulegray}\hrule height0.4pt}}}\let\@evenhead\@oddhead\def\@oddfoot{\hfil{\footnotesize\color{muted}\thepage}\hfil}\let\@evenfoot\@oddfoot}
\def\ps@opening{\ps@research\def\@oddhead{\vbox{\hbox to\textwidth{\includegraphics[width=1.10in]{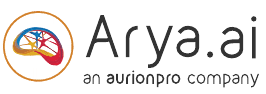}\hfill\raisebox{6pt}{\footnotesize\color{muted}23 October 2026}}\vskip7pt{\color{rulegray}\hrule height0.4pt}}}}
\makeatother
\begin{document}
\thispagestyle{opening}
\hypersetup{pdftitle={Permission to Retry: A Study of Authorization and Recovery in AI Agent Workflows},pdfauthor={Deekshith Marla; Kushagra Bhatnagar}}
\twocolumn[{\begin{@twocolumnfalse}
\noindent{\fontsize{29}{33}\selectfont\bfseries Permission to Retry:\par}
\vspace{3pt}\noindent{\raggedright\fontsize{21}{25}\selectfont\bfseries A Study of Authorization and Recovery\\in AI Agent Workflows\par}
\vspace{14pt}\noindent{\large\bfseries Deekshith Marla\qquad Kushagra Bhatnagar}\par
\vspace{3pt}\noindent{\normalsize Arya.ai}\par\vspace{15pt}
\noindent{\small\bfseries\color{arya}ABSTRACT}\par\vspace{4pt}
\noindent A durable agent can resume a consequential action while its authority changes and the action's external outcome remains uncertain. We study execution-time reliability of tool-using AI agents, isolating recovery decisions from language-model behavior. Checkpoint state, permission to dispatch, evidence of an external effect and permission to reconcile are distinct obligations. We present a finite conformance study that checks these obligations through histories scored against a provider effect ledger hidden from the recovering application. A prospectively frozen campaign executes 336 histories on native LangGraph and Temporal hosts, crossing 14 schedules, three provider contracts and four matched application policies. All 168 paired host cells have identical checked outcomes. Per host, retained-ledger reconciliation passes all 42 histories: 21 checked completions, 15 denials and six unknown outcomes with reservations retained. Cached retry, revalidation before retry and stable-key retry pass 30, 36 and 31 histories respectively. A separate 16-case component study of InfinityOS, an enterprise AI platform developed by Arya.ai, exercises exported authority, worker, recovery and settlement code over repository fakes. The results isolate the roles of current authority, provider contracts and retained observations in recovery, with safety, completion and unresolved obligations reported separately for the declared application profile.\par\vspace{8pt}
\noindent{\footnotesize\textbf{Keywords:} AI agent reliability; tool use; authorization; durable execution; recovery; uncertain outcomes.}\par\vspace{19pt}
\end{@twocolumnfalse}}]
\section{Introduction}
Consider a refund for an approved recipient and amount. The service commits the refund, but the acknowledgement disappears. Before a replacement worker resumes, the initiating principal loses write permission. Repeating the request can create a duplicate; treating a missing acknowledgement as proof of failure can release an obligation prematurely; abandoning every continuation can prevent legitimate recovery. Recovering evidence of an existing effect and obtaining authority for a new dispatch are separate decisions.

For a tool-using AI agent, resuming a workflow requires deciding what actions remain permitted and what can be concluded from earlier observations. We isolate this execution layer to examine how changing authority and uncertain tool outcomes affect recovery.

This is a distributed execution problem with established foundations. Fault-tolerant workflows already separate durable state from external effects \cite{r8}. Agent research directly addresses semantic replay, consumed authority and uncertain outcomes \cite{r1,r2,r3,r6}. These foundations motivate evaluating their composition at agent recovery boundaries. The practical question is which obligations an application preserves at each recovery boundary, under an explicit provider contract.

We study three questions. RQ1: Which configured recovery policies preserve the declared safety conditions while completing legitimate work? RQ2: Do paired applications on two independently maintained execution hosts differ under the same policy facts and provider schedules? RQ3: How does the location of a confirmed observation relative to settlement change recoverability?

Our contribution is a bounded conformance artifact and its measured application: a history specification separating admission, effects and settlement; a prospective native-host campaign; and a separately qualified InfinityOS component case study. The evaluation record preserves raw observations, exact versions and data digests, with a reconstruction checker and explicit exclusions. The policies are authored ablations, not framework defaults or independent implementations of published security protocols. Paired host outcomes are reported alongside policy differences and progress measures.

\section{Related work and contribution boundary}
AID-Guard connects authorization to provider commitment and conservative ambiguity handling \cite{r1}. CapLease conserves token-independent authorization instances and qualifies effect guarantees by the sink contract \cite{r2}. Both provide foundations for studying authorization across retries and recovery. LIMBO uses controlled tool faults and independent effects to evaluate recovery across harnesses \cite{r3}, establishing the importance of fault injection and effect-side scoring.

ACRFence identifies semantic rollback and authority resurrection, and distinguishes its validated attacks from a proposed mitigation not yet implemented in that evaluation \cite{r6}. ZeroGate separates exact admission from full lifecycle cost and makes freshness assumptions explicit \cite{r4}. Matoi separates reusable evidence, current authority and occurrence history \cite{r5}. ResidualAuth studies which authorization history must remain distinguishable under revocable delegation \cite{r7}.

We study matched application controls on native persistence/retry hosts, checked at declared authority and observation boundaries. We did not obtain a qualified licensed executable implementation of the closest security protocols from the inspected papers. Their absence is not scored as failure, and our controls are not attributed to their authors. LangGraph persistence \cite{r9} and Temporal retry semantics \cite{r10} are execution mechanisms consumed by the application, not security baselines.

\section{Model and conformance contract}
An authorization instance binds an organisation/principal scope, a canonical action, a policy epoch, expiry and an allowed effect count. It is distinct from a transport key, worker lease and checkpoint identifier. This campaign uses one synthetic refund and a one-effect allowance per history. Proposals and approvals are scripted facts, not real human attestations or model outputs.

The history contains worker claims, authority events, admissions, dispatches, provider effects, participant observations, settlements, reservation snapshots and terminal declarations. The scheduler owns provider truth and exposes only declared send/lookup responses. The checker imports neither application controls nor InfinityOS admission/recovery predicates. It reconstructs scope and freshness, approval consumption, worker ownership, effect multiplicity, retained uncertainty, settlement evidence and truthful completion.

At the initial local admission boundary, the scope, action, grant and approval must be current. For a repeated dispatch, this benchmark additionally requires current write authority and either a live stable-key deduplication contract or certified predecessor closure. This repeat-freshness rule is explicit. A protocol that permits redelivery under an already prepared grant must be judged against its own contract, not declared broken merely for differing from ours.

An already admitted first delivery can commit after local revocation. Such an effect is counted separately from an invalid local admission; no instant remote cancellation guarantee is assumed. Read-only reconciliation has a separate synthetic capability that remains valid after write revocation. It can settle an observed historical effect without authorizing another write.

The provider profiles are P1, stable-key deduplication within a declared horizon; P2, queryable outcomes without deduplication or automatic cancellation; and P3, opaque outcomes. An empty lookup cannot certify absence if an earlier request may still commit. Provider truth is never used to supply evidence that the participant did not receive. An unresolved outcome retains its unit reservation. A passing finite supplied history is not proof that all possible effects were captured. Incomplete capture or provider-contract contradictions cannot receive a passing score.

\section{Implementation and prospective study}
\subsection{Matched applications}
The application prepares an admission, dispatches and settles. Its SQLite row is separate from the host's persistence. All four controls share initial authority and action binding, reservation state and receipt-bound idempotent settlement. Cached retry sends again with a fresh key using the existing admission. Revalidation checks current write authority before a fresh-key retry and otherwise retains uncertainty. Stable-key retry preserves the original transport key. Ledger reconciliation looks up the previous delivery, settles a confirmed effect or retains uncertainty, and does not redispatch.

LangGraph 1.2.16 with SQLite checkpointer 3.1.2 runs prepare, dispatch and settle nodes. A controlled exception leaves the current node pending; the process closes its databases and exits. A second process loads the same checkpoint and application row and invokes the pending graph with its original persisted state. Completed graphs are inspected without invoking nodes again.

Temporal Python SDK 1.34.0 uses the same application through activities. The isolated CLI 1.9.1 server, embedding Temporal 1.32.0, persists to a temporary SQLite file. The scheduler kills a worker with SIGKILL before activity acknowledgement. A different process receives the native retry after the configured two-second activity timeout. The workflow run identity and retry attempt are recorded. The logical interruption positions match the graph study, while the mechanisms differ. Neither host is killed inside an active database transaction.

\subsection{Corpus and capture}
The manifest fixes two hosts, 14 schedules, three provider contracts and four policies: 336 histories, 168 per host and 42 per host-policy combination. Schedules cover completed-work reopen; revocation, policy change, expiry and payload drift before admission; pre-preparation interruption; lost acknowledgement; delayed commitment after an empty query; interruption before settlement and after settlement but before host acknowledgement; and expired deduplication. Recovery cases have unchanged-authority or revocation controls where applicable.

A pre-preparation interruption consumes no approval. After a dispatch node/activity completes, its confirmed observation is persisted by the host before settlement. Settlement stores the receipt identity with the application row so a repeated settlement can verify the same evidence without charging twice. A pending provider request commits at the final observation boundary, after the replacement has acted. Logical times specify ordering, not measured service latency.

The campaign was frozen in source \nolinkurl{b746778054e4} before collection; the artifact records its full revision. The tree was clean before and after execution. CPython was 3.13.14, its SQLite library 3.53.1 and Node 24.19.0. The artifact preserves every history, per-process state, version and reported result. Adapter errors abort collection rather than silently omit cells. Verification regrades all histories and independently reconstructs coverage, continuity and aggregates.

Earlier development work comprised 180 synthetic-control histories and 144 native LangGraph histories. Their purposes and denominators remain separate. This prospective campaign follows adapter development and does not claim unseen scenarios or statistical independence. Its complete research suite passes 64 tests with zero skips; intended assertions were observed failing before checker, native adapter and extended-boundary implementation.

\section{Native results}
\subsection{Safety and progress}
\begin{table*}[t]\centering\small\caption{Finite native conformance outcomes by authored policy}
\begin{tabularx}{\textwidth}{>{\hsize=2.0\hsize\raggedright\arraybackslash}X>{\hsize=0.8\hsize\centering\arraybackslash}X>{\hsize=0.8\hsize\centering\arraybackslash}X>{\hsize=0.8\hsize\centering\arraybackslash}X>{\hsize=0.8\hsize\centering\arraybackslash}X>{\hsize=0.8\hsize\centering\arraybackslash}X}
\toprule
\textbf{Host and authored policy} & \textbf{Passing histories} & \textbf{Reported successes} & \textbf{Checked completions} & \textbf{Denied} & \textbf{Unknown} \\
\midrule
LangGraph: Cached retry & 30 / 42 & 27 & 15 & 15 & 0 \\
LangGraph: Revalidate retry & 36 / 42 & 21 & 15 & 15 & 6 \\
LangGraph: Stable-key retry & 31 / 42 & 27 & 16 & 15 & 0 \\
LangGraph: Ledger reconciliation & 42 / 42 & 21 & 21 & 15 & 6 \\
Temporal: Cached retry & 30 / 42 & 27 & 15 & 15 & 0 \\
Temporal: Revalidate retry & 36 / 42 & 21 & 15 & 15 & 6 \\
Temporal: Stable-key retry & 31 / 42 & 27 & 16 & 15 & 0 \\
Temporal: Ledger reconciliation & 42 / 42 & 21 & 21 & 15 & 6 \\
\bottomrule\end{tabularx}
\end{table*}
Table I reports finite counts. Passing histories include correctly denied work and unresolved work; checked completion additionally requires a truthful successful terminal outcome. The 15 denials per policy concern initial invalid authority/action/expiry or revocation before preparation. They are not counted as completed tasks. No history was scored not measured in this captured corpus.

All 168 paired host cells agree on safety, progress, effect/reservation counts and finding codes. This is a null host difference for shared authored policies, not equivalence of all framework features. The experiment confirms that the observed policy distinction survives each qualified native recovery path; it does not establish that either framework defaults to the unsafe controls.

\begin{figure*}[!t]\centering\includegraphics[width=\textwidth]{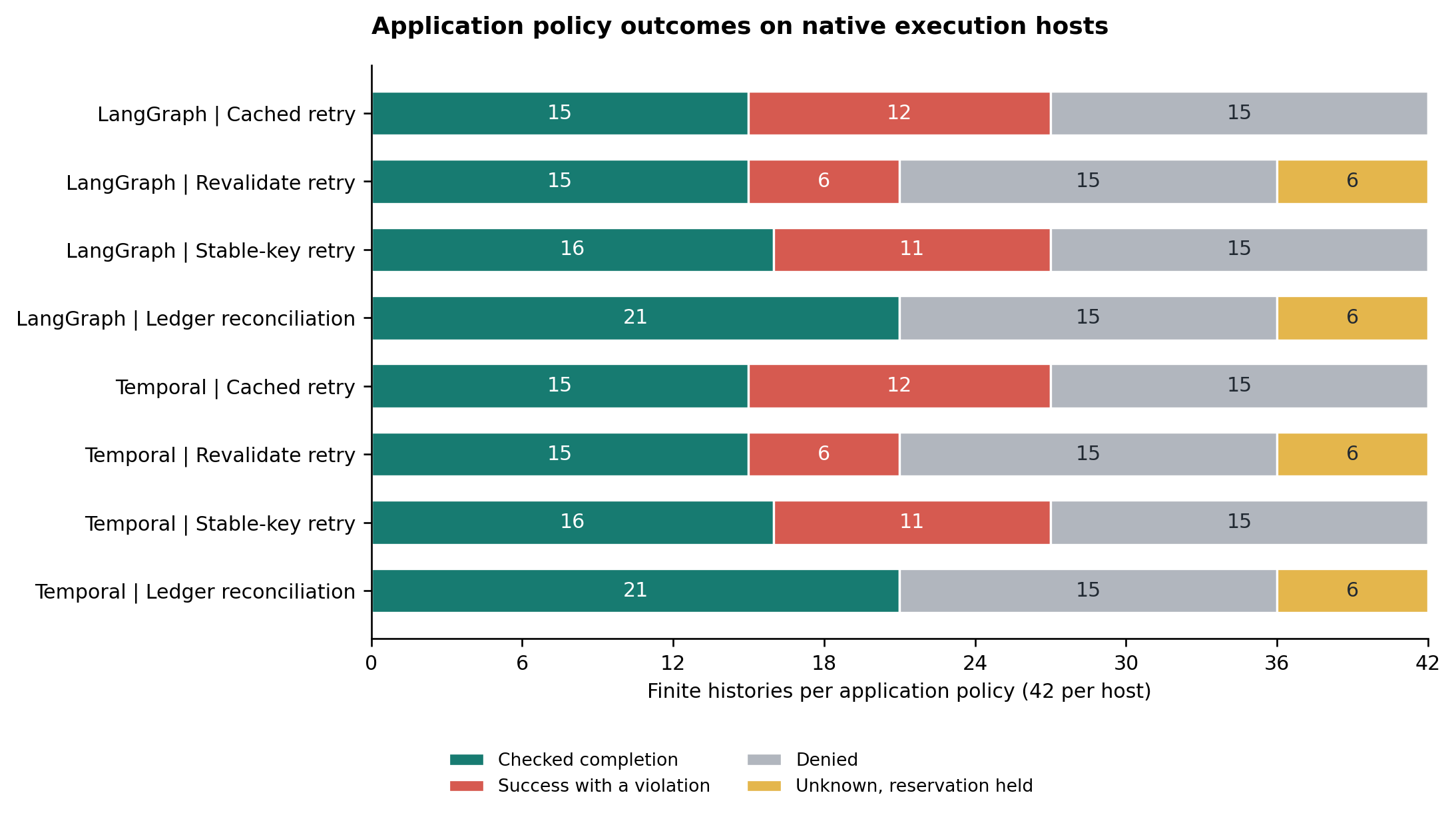}\caption{Finite policy outcomes. Each bar contains 42 histories. Checked completion, reported success with a violation, denial and unresolved work are separate.}\end{figure*}
Cached retry reports success in 27 histories per host, but only 15 are checked completions. Its 12 repeated-delivery histories exceed the effect allowance. Revalidation removes post-revocation fresh sends, but six still-authorized retries duplicate effects. It passes 36 histories while providing the same 15 checked completions as cached retry, and retains six unknown reservations.

Stable-key retry has 16 checked completions. Its contract suppresses some duplicates, but nine histories exceed the effect allowance where deduplication is unavailable or expired. Six repeat dispatches fail the benchmark's current-authority rule; these findings overlap effect violations and must not be summed as independent histories. Stable transport identity addresses a different obligation from current authority.

Ledger reconciliation passes all 42 supplied histories per host, with 21 checked completions, 15 denials and six unknown reservations. Per host, three unknowns concern opaque providers after acknowledgement loss or horizon expiry, and three concern pending commitments across the provider profiles. The exact provider/scenario cells remain visible in Figure 2 and the raw artifact. Zero findings in this finite set is not a universal guarantee.

\subsection{Why the observation boundary matters}
The provider profile alone does not determine recoverability. When a confirmed observation is already retained by the host or application, interruption before settlement or after settlement can complete without another provider write, even for P3. By contrast, P3 acknowledgement loss has no participant-visible receipt. In the pending-request cases, P1/P2 can initially return an empty lookup and still later commit; the ledger policy retains its reservation rather than manufacture certified absence. A post-revocation remote effect from that first admitted delivery remains separate from admitting a new write after revocation.

\begin{figure*}[!t]\centering\includegraphics[width=\textwidth]{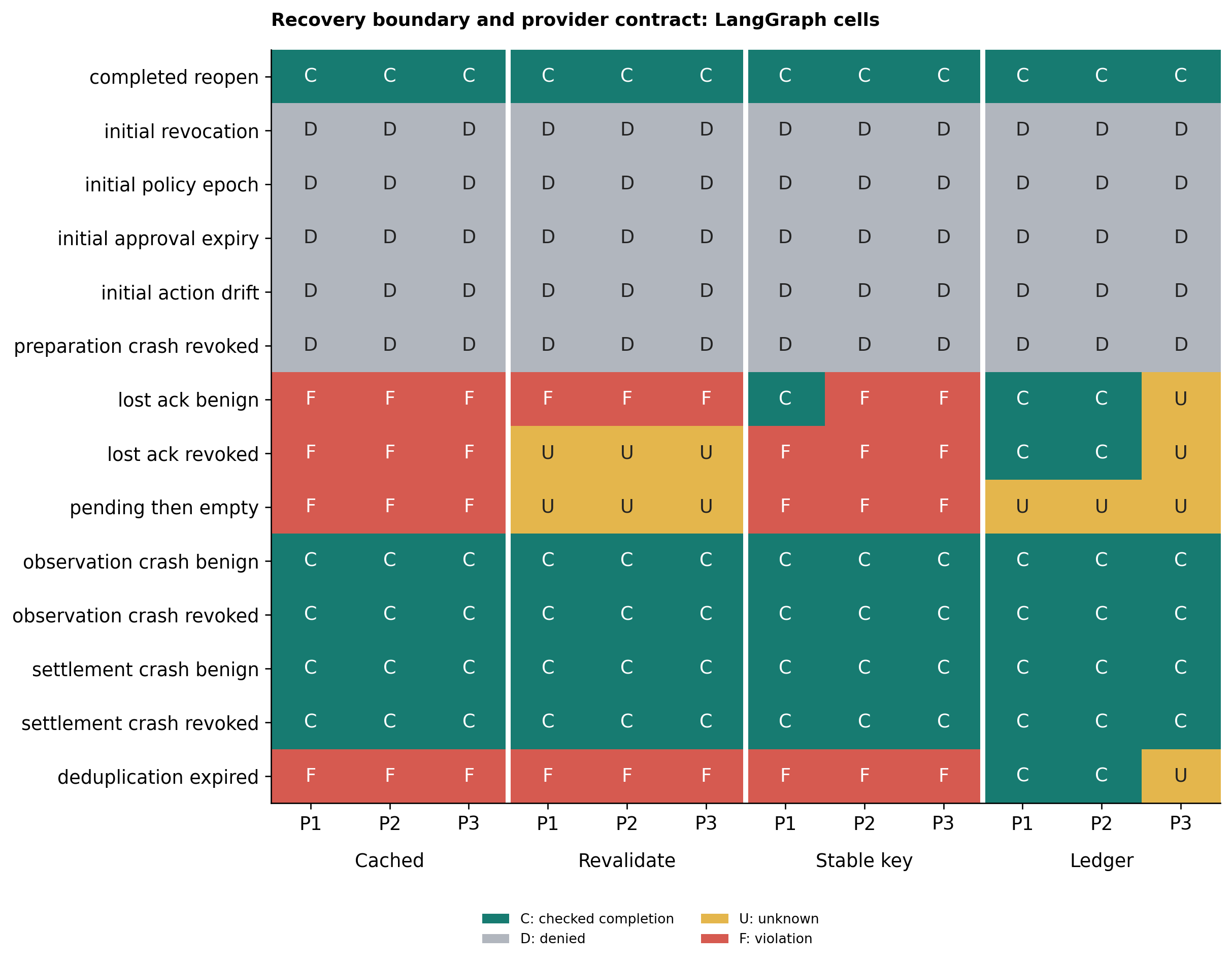}\caption{Boundary coverage for the 168 LangGraph cells. C means checked completion, D denial, U unresolved reservation and F a conformance violation. P1 is deduplicating, P2 queryable and P3 opaque. All paired Temporal cells have the same checked outcomes.}\end{figure*}
\subsection{Observed fixture execution cost}
Table II summarizes whole replacement-worker wall times. These include Python startup, IPC, scheduling and the configured timeout. Temporal's timeout/retry wait is part of the tested protocol, whereas the graph exception is already closed before replacement starts. This is neither an apples-to-apples latency ranking nor a product throughput result. The finite p95 is an empirical order statistic over heterogeneous schedules, not a population estimate or confidence interval.

\begin{table*}[t]\centering\small\caption{Whole replacement-worker fixture wall durations}
\begin{tabularx}{\textwidth}{>{\hsize=1.6\hsize\raggedright\arraybackslash}X>{\hsize=0.85\hsize\centering\arraybackslash}X>{\hsize=0.85\hsize\centering\arraybackslash}X>{\hsize=0.85\hsize\centering\arraybackslash}X>{\hsize=0.85\hsize\centering\arraybackslash}X}
\toprule
\textbf{Host} & \textbf{Observations} & \textbf{Median ms} & \textbf{Empirical p95 ms} & \textbf{Maximum ms} \\
\midrule
LangGraph & 168 & 261.4 & 316.7 & 347.3 \\
Temporal & 168 & 2149.5 & 2824.1 & 3048.8 \\
\bottomrule\end{tabularx}
\end{table*}
\section{InfinityOS component case study}
The separate case study imports InfinityOS's exported authority resolver and the existing internal Sandbox harness. Ten synthetic authority cases check current grants, trust/approval revocation, policy change, dependency refusal, lease expiry, cancellation, wrong-run scope and approval expiry. All ten return the specified grant or denial. These test dependency composition; they do not validate real identity-provider ordering or cryptographic human confirmation.

\begin{table*}[t]\centering\small\caption{Separate InfinityOS component recovery observations}
\begin{tabularx}{\textwidth}{>{\hsize=1.6\hsize\raggedright\arraybackslash}X>{\hsize=0.85\hsize\centering\arraybackslash}X>{\hsize=0.85\hsize\centering\arraybackslash}X>{\hsize=0.85\hsize\centering\arraybackslash}X>{\hsize=0.85\hsize\centering\arraybackslash}X}
\toprule
\textbf{Interruption} & \textbf{Provider profiles} & \textbf{Cases} & \textbf{Settled receipts} & \textbf{Unresolved reservations} \\
\midrule
Lost acknowledgement & P1, P2, P3 & 3 & 2 & 1 \\
Rolled-back settlement with retained observation & P1, P2, P3 & 3 & 3 & 0 \\
\bottomrule\end{tabularx}
\end{table*}
Six additional cases execute the actual action coordinator, worker, recovery and settlement code, crossing response loss and rolled-back settlement with the three provider profiles. The substrate is the existing repository fake and scripted engine. A separate synthetic provider records one effect in every case, independently of the product ledger. No case redispatches that effect. Old worker heartbeats are refused after lease replacement, without increasing effect count.

The opaque lost-acknowledgement case retains its uncertain reservation. The opaque settlement- rollback case recovers from its retained confirmed observation without provider lookup. These results support the same observation-boundary distinction as the native corpus. Separately, 17 existing action-crash and real-Weave sequential-approval regression tests pass on the pinned source. Those regressions exercise Sandbox approval/resume paths but are not new independent- provider experiments and are not pooled with either study.

The 16 new component cases do not qualify physical Mongo durability, the complete human-approval lifecycle or deployed InfinityOS behavior. The six new recovery cases use a scripted engine; the separate existing real-Weave tests do not change that experimental tier.

\section{Threats to validity}
The policies, provider and checker originate in the same project. Separate effect truth and reconstruction, hand-authored counterexamples, corrupted-artifact tests and prior RED assertions reduce some circularity; they do not substitute for independent human review or a separately implemented oracle. The common application logic also makes matched outcomes unsurprising.

One canonical action with scripted proposals does not establish generalization to arbitrary tools, model behavior, concurrent agents, delegation graphs or real authority stores. The benchmark's repeat-freshness contract is stricter than some prepared-redelivery contracts. Its results must not be used to attribute violations of guarantees those systems never offered.

The corpus is finite and deterministic. It was fixed prospectively after development, not sampled from deployments or held out from all adapter testing. There are no frequency estimates, statistical replications or acceptance-probability claims. A correct refusal can pass safety; progress metrics are reported separately to expose conservative abstention.

Native worker races, abrupt failure inside transactions, lost/corrupted databases, settlement-log capture gaps and real IAM check/commit contention remain unmeasured. Logical event order establishes fixture happens-before relationships, not atomic serialization with a third-party permission service. Python socket refusal protects the graph fixture from accidental calls; the local Temporal server/SDK are trusted participants, not an adversarial network sandbox.

No actual provider, billing account, model server, customer dataset or Production environment was used. Library and local SQLite behavior do not establish SaaS or on-premise product readiness. No independently maintained security-protocol implementation was executed. These limits constrain the contribution to conformance and replication evidence for the declared application profile.

\section{Implications and conclusion}
Within the measured profile, three questions must remain explicit at resume: is a new dispatch currently authorized; is a previous delivery still capable of committing; and is there durable participant-visible evidence that permits settlement? A checkpoint, current grant or stable key alone does not answer all three. The finite results distinguish safety, legitimate progress and conserved uncertainty without ranking framework defaults. The artifact supplies concrete histories and counterexamples for evaluating other implementations under declared contracts.

\section{Availability, ethics and AI assistance}
The accompanying publication package contains the manuscript source, figures, machine-readable outcome tables and timing observations. Original captures, manifests, implementation and detailed reproduction instructions are retained in the authors' research record; this version provides no public execution repository. The InfinityOS component case study depends on proprietary product source and repository fakes, which are excluded from the publication package. The accompanying tables permit inspection of the reported aggregates; they do not independently reproduce the experiments.

All identities, approvals and business actions are synthetic. Experiments used local fixtures; no customer data, live provider or model service was involved. Dependency and documentation retrieval occurred separately from testing. AI tools assisted with research-code development, capture analysis and manuscript preparation. The same-project origin of the controls, fixtures and checker is addressed in the threats to validity.

\appendix
\section{Reproducibility}
The manifest, original histories, host records, InfinityOS component observations, report and data digests are preserved in the authors' experiment archive. The checker independently regrades all 336 histories; the verifier checks Cartesian coverage, versions, persistence continuity and aggregate counts. The 16 component cases and 17 existing regressions have separate records. A reproduction may generate different checkpoint UUIDs and wall times while preserving its scenario definitions and checked outcomes; exact source and dependency versions must remain visible.

The archived reproduction guide contains isolated dependency setup and test, collection, verification, analysis and offline IEEE-PDF commands. Native storage is temporary and deleted after each experiment. The publication package includes derived per-history outcomes and replacement-worker timings, aggregate counts, figures and exact software versions. Full reruns require the archived implementation and dependencies; the InfinityOS component cases additionally require proprietary source. Full public reproduction is therefore not claimed for this version.

\end{document}